\documentclass{ieeeaccess}
\usepackage{cite}
\usepackage{amsmath,amssymb,amsfonts}
\usepackage{algorithmic}
\usepackage{graphicx}
\usepackage{textcomp}
\usepackage{amsmath}
\usepackage{amsfonts} 
\usepackage{algorithmic}
\usepackage{multirow}
\usepackage{array}
\usepackage{stfloats}
\usepackage{url}

\usepackage{tabularx}
\usepackage{booktabs}

\def\BibTeX{{\rm B\kern-.05em{\sc i\kern-.025em b}\kern-.08em
    T\kern-.1667em\lower.7ex\hbox{E}\kern-.125emX}}
\begin{document}
\history{Date of publication xxxx 00, 0000, date of current version xxxx 00, 0000.}
\doi{10.1109/ACCESS.2023.0322000}

\title{Spatial Attention-based Distribution Integration Network for Human Pose Estimation}
\author{\uppercase{SIHAN GAO}\authorrefmark{1},
\uppercase{JING ZHU}\authorrefmark{2}, XIAOXUAN ZHUANG\authorrefmark{3},ZHAOYUE WANG \authorrefmark{3}, and Qijin Li\authorrefmark{3}}

\address[1]{School of Mathematics and Information Science, Guangzhou University, Guangzhou 510006, China}
\address[2]{School of Electronics and Communication Engineering Lab Center, Guangzhou University, Guangzhou 510006, China}
\address[3]{School of Electronics and Communication Engineering, Guangzhou University, Guangzhou 510006, China}
\tfootnote{This work was supported by the "China National Student Innovation and Entrepreneurship Training Programme".}
\markboth
{Author \headeretal: Preparation of Papers for IEEE TRANSACTIONS and JOURNALS}
{Author \headeretal: Preparation of Papers for IEEE TRANSACTIONS and JOURNALS}

\corresp{Corresponding author: Jing Zhu (zhujing@gzhu.edu.cn}

\begin{abstract}
In recent years, human pose estimation has made significant progress through the implementation of deep learning techniques. However, these techniques still face limitations when confronted with challenging scenarios, including occlusion, diverse appearances, variations in illumination, and overlap. To cope with such drawbacks, we present the Spatial Attention-based Distribution Integration Network (SADI-NET) to improve the accuracy of localization in such situations. Our network consists of three efficient models: the receptive fortified module (RFM), spatial fusion module (SFM), and distribution learning module (DLM).
Building upon the classic HourglassNet architecture, we replace the basic block with our proposed RFM. The RFM incorporates a dilated residual block and attention mechanism to expand receptive fields while enhancing sensitivity to spatial information. In addition, the SFM incorporates multi-scale characteristics by employing both global and local attention mechanisms. Furthermore, the DLM, inspired by residual log-likelihood estimation (RLE), reconfigures a predicted heatmap using a trainable distribution weight.
For the purpose of determining the efficacy of our model, we conducted extensive experiments on the MPII and LSP benchmarks. Particularly, our model obtained a remarkable $92.10\%$ percent accuracy on the MPII test dataset, demonstrating significant improvements over existing models and establishing state-of-the-art performance.
\end{abstract}


\titlepgskip=-21pt

\maketitle

\section{Introduction}
\label{sec:introduction}

Human pose estimation is a crucial task in artificial intelligence applications, which include motion detection, human-computer communication, and pose tracking. Its primary objective is to precisely locate the joints of individuals, including the head, knees, and legs. Recent advances in human pose estimation are largely attributable to the development of deep convolutional neural networks. However, this task still faces challenges posed by diverse poses, occlusions, and overlapping body parts. To tackle these challenges, various approaches have been proposed. One notable approach is the stacked hourglass network (SHNet), originally introduced by Newell et al. \cite{Newell2016}. This network architecture, which operates across multiple stages and scales, has demonstrated the importance of feature fusion at different scales in enhancing accuracy. Building upon the hourglass model, several effective improvement networks have been proposed \cite{HUANG2023323,luo2021cascaded,lu2021feature,Xiao2020}.

\begin{figure}[p]
\centering
\includegraphics[width=0.7\linewidth]{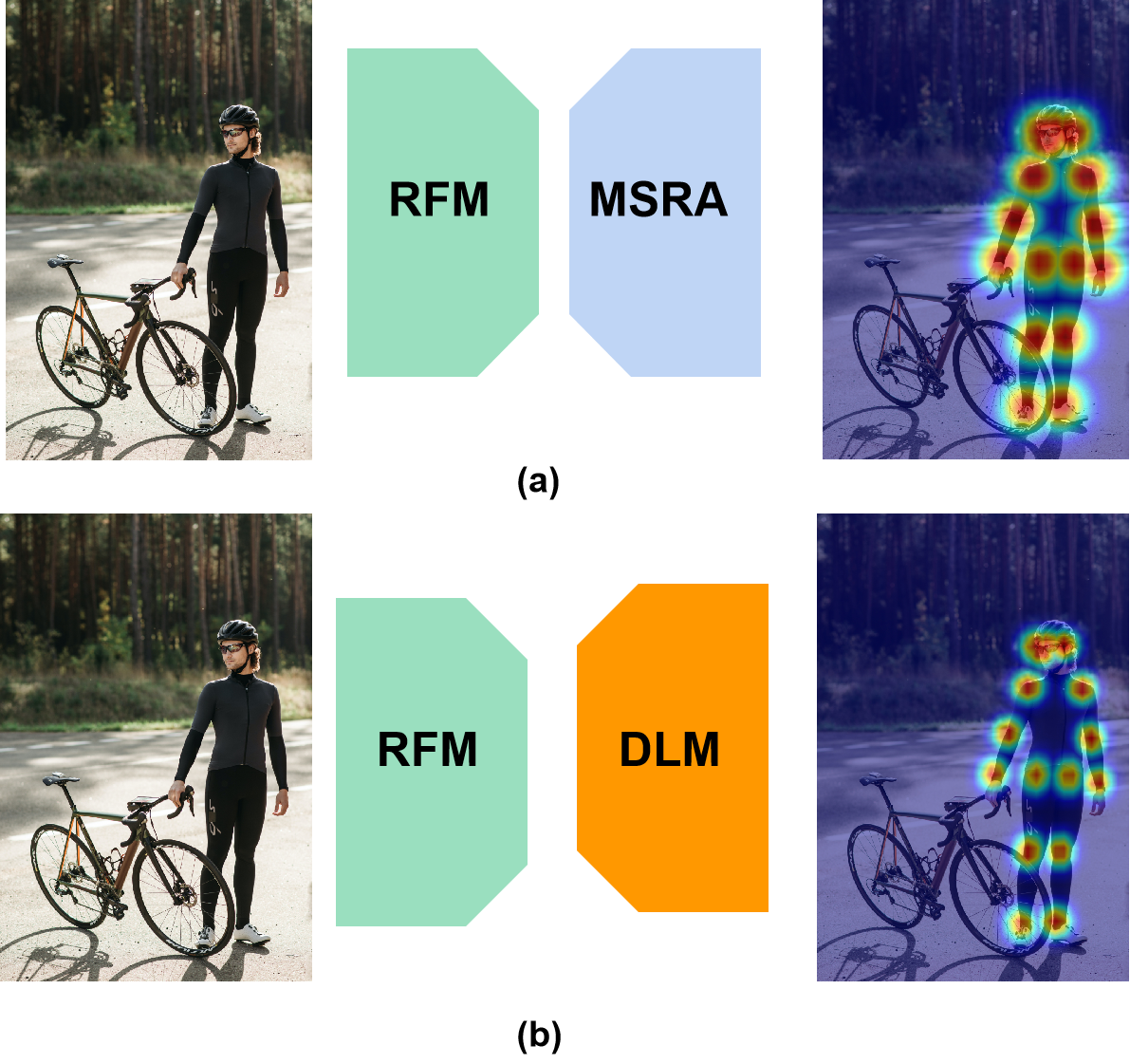}
\caption{Illustrasion of the predictions made by different distribution  strategies. (a) The heatmap of the Gaussian distribution generated a priori by the MSRA. (b) refers to the heatmap of the leanring distribution generated by the DLM. }
\label{fig:label_example}
\end{figure}

To accurately determine keypoints, human pose estimation approaches primarily utilise heatmap-based methods \cite{YU20221,cai2020learning,cheng2020higherhrnet,li2020simple,Sun2019}. A popular method for generating heatmaps was proposed by MSRA \cite{MSRA}. It is widely used in the encoding process that generates a confidence map from a Gaussian distribution to locate the keypoints. However, using this method to set up a priori distribution is challenging . In \cite{RLE}, Li et al. propose RLE, which makes the distribution learnable and facilitates the training process. It has been widely improved and rewarded by researchers in human pose estimation \cite{mcnally2022rethinking,liu2022recent,chen2022shift,mao2022poseur,huang2022capturing,li2022multi}. Furthermore, some researchers have focused on designing effective attention mechanisms such as SE channel attention\cite{SE}, CBAM channel-spatial attention \cite{CBAM} and non-local attention \cite{Zhang2019_1}. Xiao et al. \cite{Xiao2020} proposed an attention mechanism with multi-scale information.

The majority of research have concentrated on developing of plug-and-play attention modules or offered a novel multi-scale fusion technique, but they have paid relatively little attention to the reconstruction of forecast heatmaps through the process of learning the distribution of data.

In this study, we introduce SADI-NET, a spatial attention-based distribution integration network that uses a distributed learning strategy to achieve efficient estimate. Our strategy is comprised of three major components. To begin, we propose the receptive fortified module (RFM), which allows us to increase receptive fields while maintaining spatial information sensitivity. Second, we create a spatial fusion module (SFM) that uses both global and local spatial attention operations to fuse multi-stage information from RFM. Finally, a distribution learning module (DLM) is used to create a heatmap that may be trained and optimised for optimal distribution.

The following is a summary of this paper's primary contributions:

$\bullet$ We investigate a novel RFM that incorporates a dilated residual block and a plug-and-play attention mechanism. RFM units, comprising dilated convolutions, are constructed to expand the network's receptive field while maintaining spatial awareness. Additionally, squeeze-and-excitation (SE) attention mechanisms are added based on the basic block, which effectively extracts crucial information at each stage.

$\bullet$ We propose an effective distributed re-learning module, consisting of SFM and DLM. In SFM, we design a two-stage attention module that distributes the extraction of global and local information and performs scale interaction fusion. In DLM, we generate a new supervised heatmaps by learning the parameter distribution, which are effectively used in loss calculations.

$\bullet$ Our proposed network achieves state-of-the-art performance compared to current networks, as shown by experiments on the MPII and LSP datasets.

The remaining paper is organised as follows: Section 2 presents an overview of prior research on our network. SADI-Net's framework is presented in Section 3. Section 4 shows the experimental outcomes of our method. Finally, we reach our findings in Section 5.

\section{RELATED WORK}
\subsection{HUMAN POSE ESTIMATION}
Human pose estimation is a crucial issue in computer vision, and numerous solutions were presented in the literature. Traditional approaches rely on graph models with manual feature extraction techniques such as HOG and SIFT. However, these algorithms are computationally expensive, difficult to train, and susceptible to spatial details in the pictures. Recently, state-of-the-art deep learning algorithms have outperformed baseline approaches. Deeppose is able to directly regress the coordinates of keypoints through the use of deep neural networks in an efficient way. Inspired by Deeppose, scholars \cite{Tompson2014, Tompson2015, Wei2016} have proposed fully convolutional neural networks that utilize heatmaps to predict coordinates, achieving impressive results using Gaussian heatmaps. To further enhance human pose estimation performance, researchers have proposed multi-scale integration strategies \cite{Sun2019, Ou2022, Newell2016, Yang2017, Zhang2021}. Newell et al. \cite{Newell2016} introduced the Hourglass Network, a cascaded model that combines bottom-up and top-down blocks in a repetitive manner. HRNet \cite{Sun2019} employs four parallel branch networks and performs multi-stage fusion to integrate information from sub-networks. These approaches emphasize the importance of information integration on different scales to improve accuracy.

\subsection{ATTENTION MECHANISM}
The attention mechanism, which Mnih et al. \cite{Mnih2014} first used to classify images, has been widely adopted in a variety of fields, including saliency detection \cite{Kuen2016}, object recognition \cite{Xiao2015}, and image captioning \cite{Xu2015}. In the field of human pose estimation, several studies \cite{Tian2019, Liu2018} have employed different strategies to incorporate the attention mechanism into their models. Chu et al. \cite{Chu2017} effectively utilized attention mechanisms by redesigning the Hourglass Network and integrating visual attention blocks, resulting in improved keypoint localization using heatmap-based models. Furthermore, Woo et al. \cite{CBAM} proposed CBAM, comprising channel attention and spatial attention components, and demonstrated its effectiveness.

\section{PROPOSED METHOD}
\begin{figure*}[h]
\centering
\includegraphics[width=\linewidth]{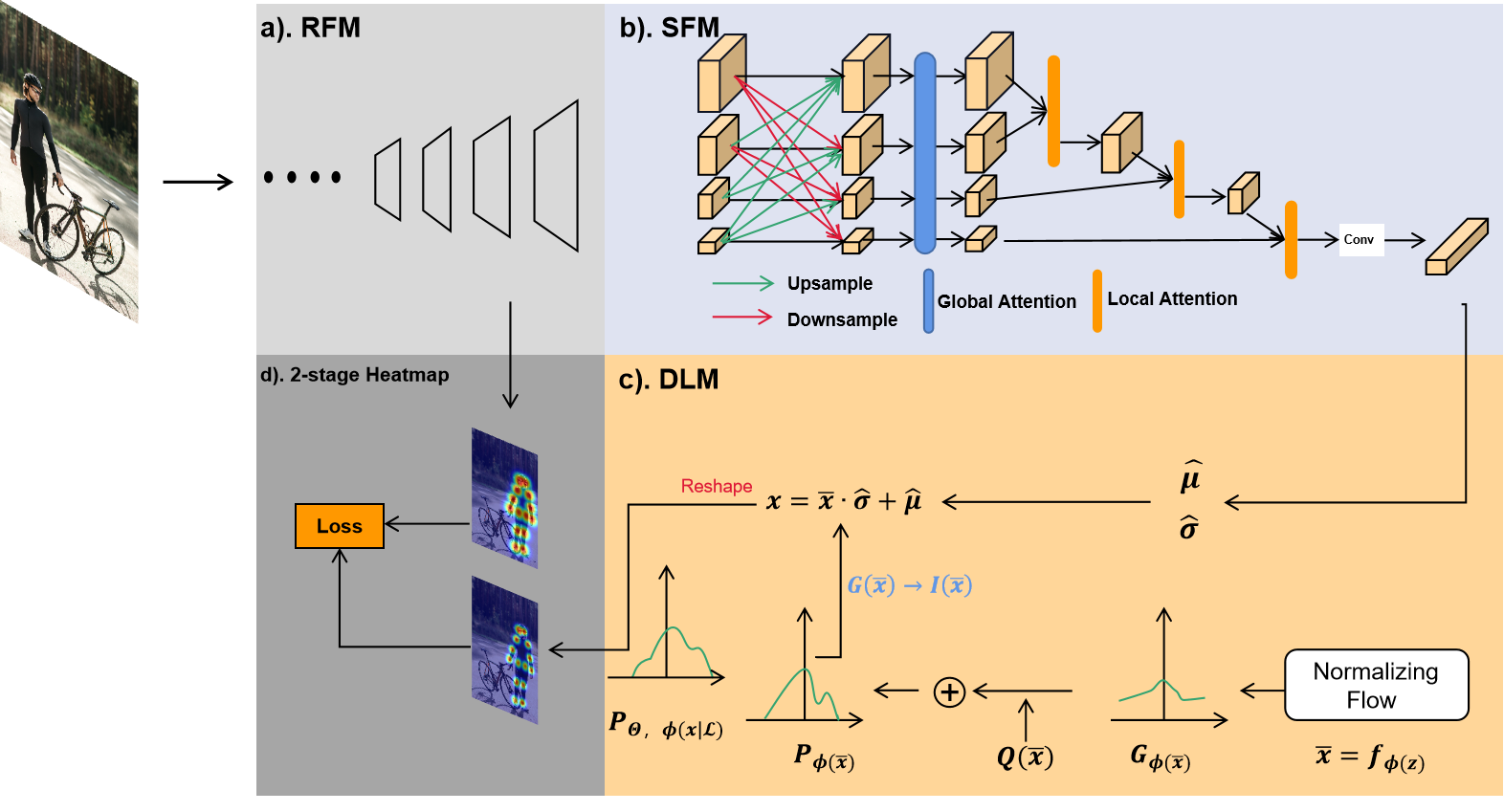}
\caption{The architecture of SADI-Net. (a) Abbreviated RFM. (b) The DLM obtains multi-scale information from (a) RFM. Then, the SFM contextually combines multi-stage features under multi-stage with global and local attention mechanisms. (c) The DLM (based on RLE \cite{RLE}) generates a heatmap. (d) Heatmaps from a and c.}
\label{fig:model}
\end{figure*}

In this section, first, we review the classic hourglass network in \ref{3.1}. Then, in \ref{3.2}, we introduce the RFM, a new hourglass unit comprising additional dilated convolution and SE attention mechanisms. In \ref{3.3}, the SFM is proposed, which acquires multi-scale features and generates a classification output using multi-scale spatial attention mechanisms. Finally, we present the DLM in \ref{3.4}, which can reconstruct heatmaps in the training process.

\subsection{Revisiting the SHNet}
\label{3.1}
The SHNet is a highly regarded architecture for estimating human poses, primarily owing to its exceptional ability in balancing the requirement for both a large receptive field and high resolution. It adopts a top-down approach, where an input image is first processed through residual blocks to reduce its resolution, generating low-resolution feature maps. These feature maps are then increased in resolution and combined with the original, high-resolution feature, leading to the production of heatmaps that precisely predict the position of keypoints.

\subsection{Receptive Fortified Module}
\label{3.2}
The majority of deep learning networks rely on residual networks and use bottlenecks or fundamental blocks. In order to enhance keypoint detection, the encoding process often involves repeated down-sampling operations, which expand the receptive field while reducing the number of feature map elements. However, this down-sampling can result in decreased spatial resolution, potentially leading to the loss of crucial local details necessary for accurate estimation. Drawing inspiration from dilated convolutions and SE modules, we introduce a novel Receptive Fortified Module (RFM), to maintain the network's ability to capture local feature information while enlarging the receptive fields.

\begin{figure}[h]
\centering
\includegraphics[width=0.6\linewidth]{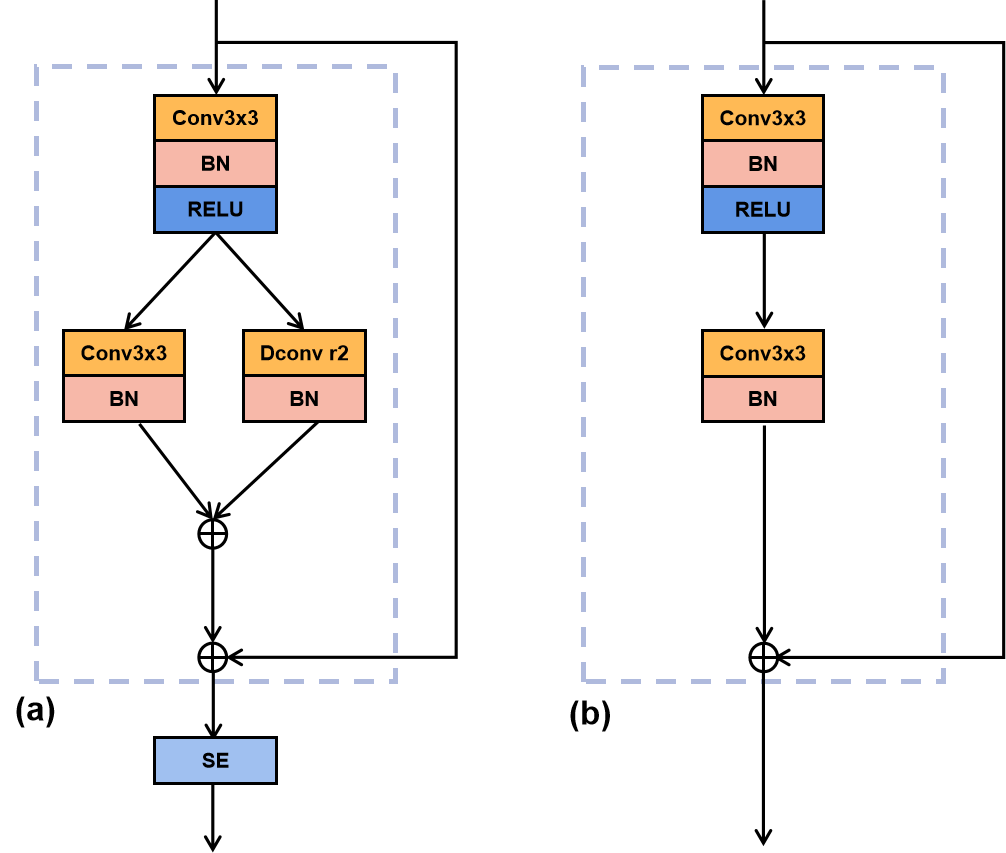}
\caption{(a)Dilated residual block. (b)Basicblock}
\label{fig:d}
\end{figure}

In the HGNet architecture, the basicblock structure comprises two consecutive $3\times 3$ convolutional layers connected by an identity mapping acting as a shortcut connection. In our approach, we introduce the replacement of the basicblock with a dilated residual block, as illustrated in Figure \ref{fig:d}. This modified block includes an additional parallel dilated convolutional layer, in addition to the $3\times 3$ convolutional layers of the basicblock. The output is then obtained from the SE block. The dilated convolution is mathematically defined as:

\begin{equation}
f_{out}\left[ i,j \right] =\sum_{m=0}^{k-1}{\sum_{n=0}^{k-1}{\left( f_{inp}\left[ i+b*m,j+b*n \right] *w\left[ m,n \right] \right)}}
\end{equation}

Here, $f_{out}$ represents the output information at index $[i,j]$, and $f_{inp}$ corresponds to the corresponding input feature map. An index indicating the location of the dilated convolutional kernel is denoted by $\left[p,q\right]$, where $w$ represents the kernel size and $b$ denotes the dilation rate. In our proposed approach, the backbone model is constructed using RFM units, which consist of dilated residual blocks. The predicted heatmaps generated by the backbone network are denoted as $H_1$.
\subsection{Spatial Fusion Module}
\label{3.3}

In the conventional classification methodology, high-resolution feature maps undergo a down-sampling process and are fused multiple times to generate the final classification output. However, this down-sampling procedure leads to a decrease in feature map resolution, which comprises a higher degree of contextual semantics but a lower level of spatial detail. This reduction in spatial detail diminishes the model's ability to accurately locate keypoints. Attention mechanisms are widely utilised to address this limitation and enhance spatial details, but plug-and-play mechanisms, such as the SE \cite{SE}, CMBA \cite{CBAM} and ECA \cite{ECA}, do not cater specifically to human pose estimation models. To overcome this limitation, SPCNet proposes the selective information module, which accumulates multiple features and allows for the adaptive fusion of information. However, it only employs one-stage fusion strategies. TIn order to solve this problem, we have developed the SFM that is depicted in Figure \ref{fig:model}. This model is responsible for producing the final classification result. Specifically, first, the multi-scale feature(i.e., 8, 16, 32 and 64 pixels) from the RFM are down- or up-sampled several times to the same resolution and then fused.The global attention mechanism enables the adaptation of feature maps, where fusion is carried out on the basis of the global spatial weights and subsequently output. subsequent, the high-resolution and low-resolution features undergo multiple-stage fusion through the local attention mechanism and produce a classification output.

\vspace{4mm}
\begin{figure}[t]
	\centering
	\begin{minipage}{\linewidth}
		\centering
		\includegraphics[width=0.9\linewidth]{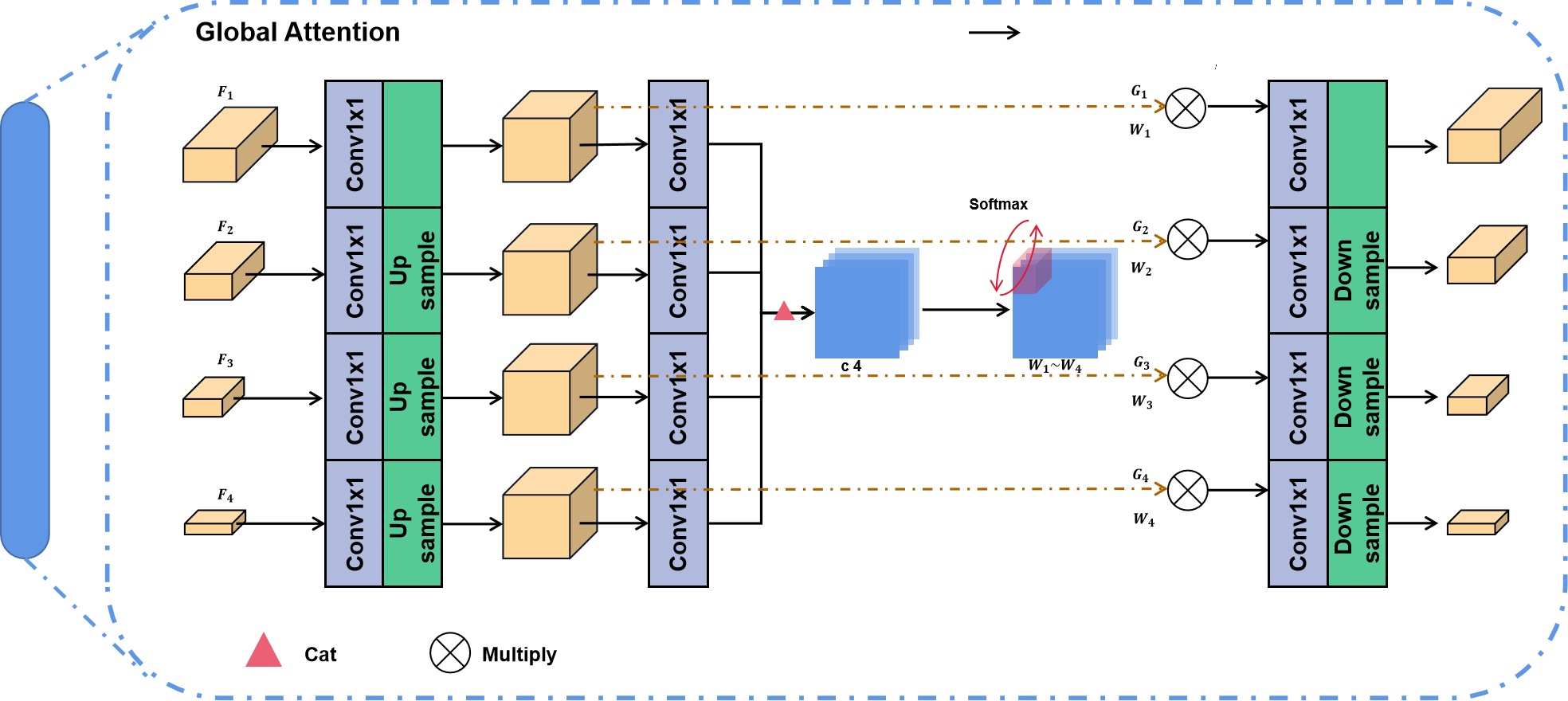}
		\caption{Global attention mechanism}
		\label{fig:global}%
	\end{minipage}
    \vspace{4mm}

	\begin{minipage}{\linewidth}
		\centering
		\includegraphics[width=0.9\linewidth]{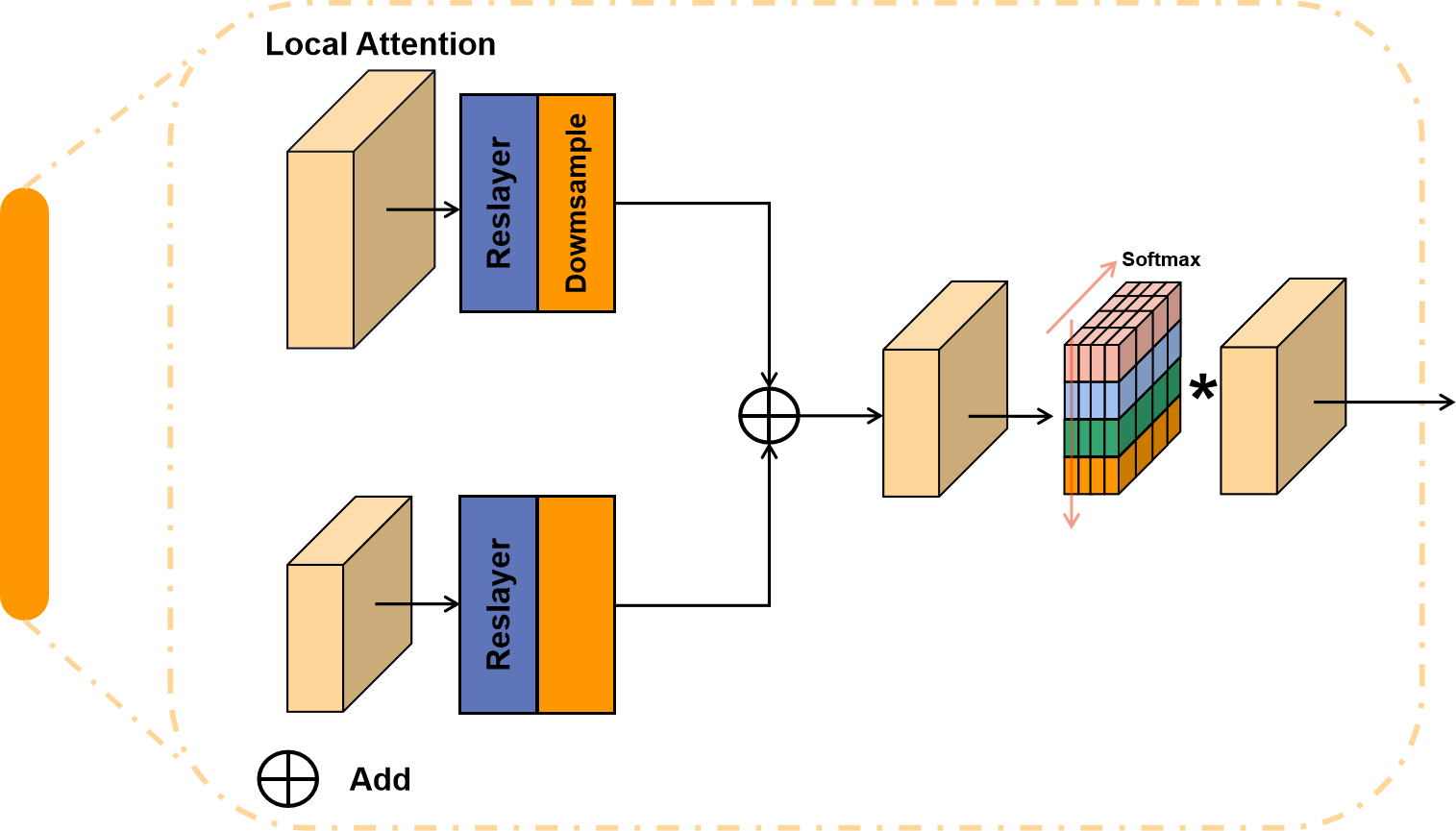}
		\caption{Local attention mechanism}
		\label{fig:Local}%
	\end{minipage}
\end{figure}

\textbf{Global attention mechanism.} As illustrated in Figure \ref{fig:global}, we adjust the channels of the multiple fusion features, denoted as $\mathrm{F}_1\sim \mathrm{F}_4$, which have 64, 128, 256, and 512 channels, respectively, to 256 channels using 1$\times$1 convolutional layers. Subsequently, we up-sample the multi-level features to a size of 64$\times$64, resulting in features denoted as $G_1\sim G_4$. To achieve this, we employ $1\times 1$ convolutional filters to project the multiple levels of feature maps into four single-channel features, which are then concatenated as $\mathrm{U} = \mathrm{U}_1\sim \mathrm{U}_4$, corresponding to $\mathrm{G}_1\sim \mathrm{G}_4$. To obtain trainable global weights $W$, we apply a softmax function to $\mathrm{U}$ along the channel dimension, formulated as follows:

\begin{equation}
\mathrm{W}_{\mathrm{n}}^{\mathrm{i},\mathrm{j}}=\frac{\exp \left( \mathrm{U}_{\mathrm{n}}^{\mathrm{i},\mathrm{j}} \right)}{\sum_{\mathrm{k}=1}^4{\exp \left( \mathrm{U}_{\mathrm{n}}^{\mathrm{i},\mathrm{j}} \right)}}
\end{equation}
\begin{equation}
    \sum_{\mathrm{n}=1}^4{\mathrm{W}_{\mathrm{n}}^{\mathrm{i},\mathrm{j}}=1}, 
\end{equation}

Here, $\mathrm{U}_{\mathrm{n}}^{\mathrm{i},\mathrm{j}}$ is the index of the referring single-channel feature map element. The trainable weight parameter corresponding to $\mathrm{U}{\mathrm{n}}^{\mathrm{i},\mathrm{j}}$ is denoted as $\mathrm{W}_{\mathrm{n}}^{\mathrm{i},\mathrm{j}}$. Next, the global learning weights $W_n$ and $G_n$ are multiplied, and the resulting product is resized using a $1\times 1$ convolution layer to match the size of $F_n$. This process allows the final fused features to gather multi-level information from multiple scales, enhancing the model's global information awareness capabilities.

\textbf{Local attention mechanism.} In the feature fusion phase of the classification head, we incorporate local attention mechanisms, as illustrated in Figure \ref{fig:Local}. We input two-scale feature maps and down-sample the high-resolution representation to merge it with the second-high-resolution feature. Local weighting feature maps are generated by applying a softmax function along the columns of the fused features. The final feature output is obtained by element-wise multiplying the weighting map by the fused features. Through this strategy, the information from adjacent scales are fused so that the output information can represent features from more scales.

\textbf{Classification head.}
Our classification head design is based on the HRNet classification approach. The feature maps produced by the global attention mechanism, with four different resolutions, have their channel numbers increased to 128, 256, 512, and 1024, respectively. To down-sample the high-resolution representations, we employ a 3x3 convolution with a stride of 2, resulting in an output of 256 channels. These representations are then added to the second-highest resolution representations, which are repeated twice to generate 1024-channel features and further increase the channel count.

\subsection{Distribution Learning Module}
\label{3.4}
The ground-truth heatmaps encoded with Gaussian kernels not only demonstrate spatial generalisation capabilities but also enhance the interconnectedness of key points. Due to the unpredictable nature of the underlying distribution, employing a Gaussian distribution for the purpose of generating heatmaps cannot facilitate a change in the distribution during the training process. RLE \cite{RLE} first proposed an approach to learning distribution changes. Specifically, it defines a loss function based on residual log-likelihood estimation, which can be described as follows:
\begin{align}
  \mathcal{L} _{rle} &= -\log P_{\vartheta ,\xi }\left( x|\mathcal{T} \right) |_{\mathbf{x}=\boldsymbol{\mu }_g}
 \\
    &= -\log Q\left( \bar{\mu}_g \right) -\log G_{\xi }\left( \bar{\mu}_g \right) -\log s+\log \hat{\sigma} \nonumber
\end{align}
The expression $P_{\vartheta ,\xi }\left( x|\mathcal{T} \right)$ denotes the probability of the ground truth being present at location $x$ given the input image $\mathcal{T}$. In this context, $\vartheta$ represents the parameters of SFM, while $\xi $ corresponds to the parameters of Real NVP, a flow-based model. The distribution $Q\left( \bar{\mu}_g \right) $ follows a Gaussian distribution, and $G_{\xi}\left( \bar{\mu}_g \right) $ denotes the distribution learned by the Real NVP. 

Although the previous method has been effective, it is based solely on regression and does not utilize heatmap-based techniques. To circumvent this constraint, we propose a DLM, as depicted in Figure \ref{fig:model}, which generates adaptive heatmaps through a probability distribution. Drawing inspiration from RLE \cite{RLE}, we compute the probability distribution $P_{\vartheta ,\xi }\left( x|\mathcal{T} \right)$ to indicate the likelihood of the ground-truth keypoint being present at location $x$ given the input image $\mathcal{T}$.

Our approach differs from RLE in how we utilize the flow model $f_{\xi }$ to capture the underlying distribution of predicted heatmaps. Instead of solely mapping an initial $\bar{z}\sim \mathcal{N} \left( 0,1 \right)$ to a complex distribution $\bar{x}\sim G_{\xi }\left( \bar{x} \right)$ and comparing it to the ground truth, we incorporate a transformation step to enhance the smoothness of the distribution. This transformation is expressed as $\bar{x}\sim I_{\xi }\left( \bar{x} \right)$.

To accomplish this, we apply the subsequent formula to the input image $\mathcal{T}$:
\begin{equation}
I_{\xi }\left( \bar{x} \right) =G_{\xi }\left( \frac{\bar{x}-\mu}{\sigma} \right)
\end{equation}

Here, $\mu$ and $\sigma$ represent the mean and standard deviation of the distribution $G_{\xi }\left( \bar{x} \right)$, respectively. We then obtain $P_{\xi }\left( \bar{x} \right)$ by adding a standard gaussian distribution $Q\left( \bar{x} \right)$ to $I\left( \bar{x} \right)$. The distribution $P_{\vartheta ,\varPhi}\left( x|\mathcal{L} \right)$ is built upon $P_{\xi }\left( \bar{x} \right)$, where $x=\bar{x}\cdot \hat{\sigma}+\hat{\mu}$. Finally, we shift and rescale the distribution $P_{\vartheta ,\varPhi}\left( x|\mathcal{L} \right)$ to generate the learnable heatmaps $H_2$, which have the same size as the predicted heatmaps $H_1$ obtained from the backbone.

\subsection{Loss Functions}
 \label{3.5}
The MSE loss function is commonly employed in human pose estimation. Building upon MSE, we have redefined the loss function to jointly predict score maps using both $H_1$ and $H_2$ heatmaps. The loss function for RFM predictions can be defined as follows:

\begin{equation}
\mathcal{L} _{mse}^{H_{\mathbf{1}}}=-\log P_{\vartheta _f}\left( \mathrm{x}|\mathcal{T} \right) |_{\mathbf{x}=\boldsymbol{\mu }_{\boldsymbol{g}}}
\end{equation}

Here, $\vartheta _f$ represents a parameter of the RFM. Similarly, the loss for the distribution learning model ($H_2$) can be defined as:

\begin{equation}
\mathcal{L} _{mse}^{H_2}=-\log P_{\xi  _{\mathbf{q}}}\left( \mathrm{x}|\mathcal{T} \right) |_{\mathbf{x}=\boldsymbol{\mu }_{\boldsymbol{g}}}
\end{equation}

In this case, $\xi  _{\mathbf{q}}$ denotes a parameter of the DFM. Finally, the total loss is calculated by combining the two loss functions:

\begin{equation}
\mathcal{L} =\left( 1-\chi \right) \mathcal{L} {mse}^{H{\mathbf{1}}}+\chi \mathcal{L} {mse}^{H{\mathbf{2}}}
\end{equation}

Here, $\chi$ is a constant used to balance the two losses. By default, we set $\chi = \sigma$.

\section{EXPERIMENTS AND RESULT ANALYSIS}

\begin{table*}[h]
  \caption{Results obtained using the MPII test dataset(PCKh@0.5).}
  \label{tab1}
  \centering
  \renewcommand{\arraystretch}{1.5} 
  \begin{tabularx}{\textwidth}{p{2.5cm}XXXXXXXX}
    \hline
    \textbf{Methods} & \textbf{Head} & \textbf{Shoulder} & \textbf{Elbow} & \textbf{Wrist} & \textbf{Hip} & \textbf{Knee} & \textbf{Ankle} & \textbf{Mean} \\
    \hline
    \multirow[m]{1}{*}{Tompson et al.\cite{Tompson2014}   } & 95.8 & 90.3 &80.5 & 74.3 &77.6 &69.7&62.8 &79.6 \\
                       
    \multirow[m]{1}{*}{Pishchulin et al.\cite{Pishchulin2016}}    &94.1 &90.2 & 83.4 &77.3 &82.6 &76.7 &68.6 &82.4\\

    \multirow[m]{1}{*}{Rafi et al.\cite{Rafi2016}}    &97.2 &93.9 &86.4 &81.3 &86.6 &80.6 &73.4 &86.3\\
                      
    \multirow[m]{1}{*}{Insafutdinov et. al.\cite{Insafutdinov2016}}    &96.8 &95.2 &89.3 &84.4 &88.4 &83.4 &78.0 &88.5\\
    \multirow[m]{1}{*}{Wei et al.\cite{Wei2016}}    &97.8 &95.0 &88.7 &84.0 &88.4 &82.8 &79.4 &88.5\\
    \multirow[m]{1}{*}{Bulat et al.\cite{Bulat2016}}    &97.9 &95.1 &89.9 &85.3 &89.4 &85.7 &81.7 &89.7 \\
    \multirow[m]{1}{*}{Newell et al.\cite{Newell2016}}    &98.2 &96.3 &91.2 &87.1 &90.1 &87.4 &83.6 &90.9\\
    \multirow[m]{1}{*}{Chen et al.\cite{Chen2018}}    &98.1 &96.5 &92.5 &88.5 &90.2 &89.6 &86.0 &91.9\\
    \multirow[m]{1}{*}{HRNet \cite{Sun2019}}    &98.6 &96.9 &92.8 &89.0 &91.5 &89.0 &85.7 &92.3\\
    \midrule
    \multirow[m]{1}{*}{\textbf{SADI-Net}*}    &98.3 &\textbf{97.0} &92.6 &\textbf{89.1} &90.2 &88.9 &\textbf{86.2} &92.1\\
    \bottomrule
  \end{tabularx}
\end{table*}

\begin{table*}[h]
  \caption{Results obtained using the LSP test dataset(PCK@0.2).}
  \label{tab2}
  \centering
  \renewcommand{\arraystretch}{1.5} 
  \begin{tabularx}{\textwidth}{p{2.5cm}XXXXXXXX}
    \hline
    \textbf{Methods} & \textbf{Head} & \textbf{Shoulder} & \textbf{Elbow} & \textbf{Wrist} & \textbf{Hip} & \textbf{Knee} & \textbf{Ankle} & \textbf{Mean} \\
    \hline
    \multirow[m]{1}{*}{Tompson et al.\cite{Tompson2014}   } & 90.6 & 79.2 &67.9 & 63.4 &69.5 &71.0 &64.2 &72.6 \\
                       
    \multirow[m]{1}{*}{Pishchulin et al.\cite{Pishchulin2016}}    &97.0 &91.0 & 83.8 &78.1 &91.0 &82.0 &87.1 &82.4\\
    \multirow[m]{1}{*}{Insafutdinov et. al.\cite{Insafutdinov2016}}    &95.8 &86.2 &79.3 &75.0 &86.6 &83.9 &79.8 &83.8\\     
    
    \multirow[m]{1}{*}{Wei et al.\cite{Wei2016}}    &97.8 &92.5 &87.0 &83.9 &91.5 &90.8 &89.9 &90.5\\
    \multirow[m]{1}{*}{Bulat et al.\cite{Bulat2016}}    &97.2 &92.1 &88.1 &85.2 &92.2 &91.4 &88.7 &90.7 \\
    \multirow[m]{1}{*}{Zhang et al.\cite{Zhang2019}}    &97.3 &92.3 &86.8 +&84.2 &91.9 &92.2 &90.9 &90.9\\
    \multirow[m]{1}{*}{Lu et al.\cite{lu2021feature}}    &97.6 &94.2 &89.0 &83.8 &96.3 &94.1 &90.8 &92.2\\
    \midrule
    \multirow[m]{1}{*}{\textbf{SADI-Net}*}    &98.2 &95.7 &93.1 &90.8 &96.4 &95.7 &94.1 &94.9\\
	\bottomrule
  \end{tabularx}
\end{table*}

\label{sec:guidelines}
\subsection{Datasets}
In this paper, we evaluate the efficacy of our network utilising two publicly accessible human pose estimation datasets: MPII \cite{MPII} and LSP \cite{LSP}, including the extended training dataset of LSP.

\textbf{MPII.} The MPII dataset, sourced from YouTube, comprises 25,000 images and 40,000 annotated person samples, each containing 16 body joints. In line with the Hourglass Network,In the MPII dataset, we use 28,000 annotated items to train models and retain 12,000 samples for testing.

\textbf{LSP.} The LSP, along with its expanded training dataset, consists of 12,000 images of individuals participating in a variety of sports. These images come with annotations of 14 body joints. We distribute 11,000 images for training and keep 1,000 images for testing.

\subsection{Evaluation Metrics}
In our evaluation, we adopt PCKh@0.5 for MPII and PCK@0.2 for LSP as the evaluation metrics. PCK evaluates the precision of keypoint detection by calculating the fraction of detections that lie within a normalised distance of the ground truth. PCKh extends PCK by incorporating a fraction of the head size as the normalization factor. The precise formula for PCK is defined as follows:
\begin{equation}
    PCK_{i}^{k}=\frac{\sum{_p\delta \left( \frac{\boldsymbol{d}_{\boldsymbol{pi}}}{\boldsymbol{d}_{\boldsymbol{p}}^{\boldsymbol{def}}} \right) \leqslant T_{\boldsymbol{k}}}}{\sum{_p1}}
\end{equation}Here, $T_{\boldsymbol{k}}$ represents the matching threshold, and the parameter $d_{pi}$ represents the Euclidean distance between the true value of the ith keypoint and the value predicted for the pth individual., and $d_{p}$ serves as the normalization factor for the pth person, obtained by dividing the former. For PCKh@0.5, we utilize a threshold of $50\%$ of the head segment as the normalization factor. Similarly, for LSP, PCK@0.2 adopts a threshold of $20\%$ of the torso size for calculating the metric.

\subsection{Implementation details} 
The experiment in this paper was conducted using the CUDA 11.6 version and Python 3.7 version on the Ubuntu 20.04.5 LTS system with an Intel i9 processor and two NVIDIA GeForce RTX 3090 GPUs for deep learning
training. 

For training our SADI-Net, we employ Pytorch 1.12.0 version, and optimize it using the Adam algorithm. The learning rate is decreased by a factor of 10 at the 170th and 200th epochs from the original value of 0.001. To enhance the data, we incorporate random rotation, scaling, and flipping as augmentation techniques. During testing, we utilize image pyramids with six scales and flipping to generate the results. Specifically, we set the rotation range between $-80$ and $80$ degrees, the scale range between $0.5$ and $1.5$, and a flipping probability of 0.5.

\subsection{Results}
\subsubsection{Results obtained using the MPII Dataset}

Table \ref{tab1} presents the results for the MPII dataset with a threshold of 0.5. The obtained experimental results demonstrate that our methods achieved PCKh@0.5 scores of $92.1\%$. Importantly, when compared to the Hourglass Network, our approach exhibited a notable improvement of $4.0\%$ across all body joints. Additionally, our model surpassed HRNet\cite{Sun2019} in terms of performance on the shoulder, wrist, and ankle keypoints.

\subsubsection{Results obtained using the LSP Dataset}

Table \ref{tab2} shows the performance of SADI-NET and the previous models on the LSP dataset. Following the approach of previous methods\cite{Xiao2020}, we trained our network by incorporating the MPII training set into both the LSP dataset and its extended training set. Our method significantly outperformed the previous state-of-the-art results by a considerable margin of $1.4\%$. Notably, we observed substantial improvements of $2.9\%$, $6.9\%$, and $2.0\%$ for challenging keypoints such as the elbow, wrist, and ankle, respectively.

\subsection{Ablation Study}
\subsubsection{Receptive Fortified Module}
To validate the influence of the RFM on the experimental results, we conducted an experiment comparing the Hourglass Network, using a basicblock, to the RFM, of which uses a dilated residual block. Our findings, presented in Table \ref{tab3}, demonstrate that the RFM outperforms the baseline, resulting in an increase of PCKh@0.5 scores from $88.9\%$ to $89.9\%$. Therefore, our results indicate that the RFM is advantageous in effectively locating joints.

\begin{table*}[t]
  \caption{Comparison between the classic HourglassNet and the RFM on the MPII validation dataset (PCKh@0.5).}
  \label{tab3}
  \centering
  \renewcommand{\arraystretch}{1.5} 
  \begin{tabularx}{\textwidth}{p{2.5cm}XXXXXXXX}
    \hline
    \textbf{Methods} & \textbf{Head} & \textbf{Shoulder} & \textbf{Elbow} & \textbf{Wrist} & \textbf{Hip} & \textbf{Knee} & \textbf{Ankle} & \textbf{Mean} \\
    \hline
    Hourglass	& 96.7 & 95.8 &89.9 &84.9 &88.8 &85.0 &80.6 &88.9\\
    RFM		& 96.8 & 96.1 &90.4 &85.2 &89.6 &86.6 &82.1 &89.6\\
	\bottomrule
  \end{tabularx}
\end{table*}

\begin{table*}[h]
  \caption{Effect of the SPM$-$DFM performance on the MPII validation dataset(PCKh@0.5).}
  \label{tab4}
  \centering
  \renewcommand{\arraystretch}{1.5} 
  \begin{tabularx}{\textwidth}{p{2.5cm}XXXXXXXX}
    \hline
    \textbf{Methods} & \textbf{Head} & \textbf{Shoulder} & \textbf{Elbow} & \textbf{Wrist} & \textbf{Hip} & \textbf{Knee} & \textbf{Ankle} & \textbf{Mean} \\
    \hline
Hourglass	& 96.7 & 95.8 &89.9 &84.9 &88.8 &85.0 &80.6 &88.9\\
Hourglass+SPM$-$DFM	& 96.8 & 95.6 &90.3 &86.2 &89.6 &86.4 &82.4 &89.7\\
	\bottomrule
  \end{tabularx}
\end{table*}

\begin{table}[h]
  \caption{Effect of the improvement loss function on MPII.}
  \label{tab5}
  \centering
  \begin{tabular}{cc}
    \hline
    \textbf{Loss} & \textbf{Mean} \\
    \hline
    $\mathcal{L} _{mse}^{H_1}+\mathcal{L} _{mse}^{H_2}$	&89.3\\
    $\left( 1-\chi \right) \mathcal{L} _{mse}^{H_1}+\chi \mathcal{L} _{mse}^{H_2}$	 &90.1\\
    \hline
  \end{tabular}
\end{table}

\subsubsection{Distribution Learning Module}
As the primary purpose of the SFM is to provide output to the DLM, we comprehensively tested of both the modules together, which is denoted as SPM$-$DFM. Table \ref{tab4} shows that the SPM$-$DFM model obtains an increase of all keypoints joints and the mean is improved from $88.9\%$ to $90.1\%$.

\subsubsection{Comparison of different loss functions}
As mentioned in Section \ref{3.5}, we propose a new loss function that combines the outputs of RFM and DLM using the parameter $\chi$. We experimented with different combinations of MLE loss functions, including $\mathcal{L} = \chi \mathcal{L} _{mse}^{H_1}+\left( 1-\chi \right) \mathcal{L} _{mse}^{H_2}$ and $\mathcal{L} = \left( 1-\chi \right) \mathcal{L} _{mse}^{H_1}+\chi \mathcal{L} _{mse}^{H_2}$. We observed that $\mathcal{L} = \left( 1-\hat{\sigma} \right) \mathcal{L} _{mle;H_1}+\hat{\sigma}\mathcal{L} _{mle;H_2}$ effectively improved the network's performance. However, using $\mathcal{L} = \chi \mathcal{L} _{mse}^{H_1}+\left( 1-\chi \right) \mathcal{L} _{mse}^{H_2}$ could lead to learning errors, resulting in a sharp drop in accuracy during random epochs. Therefore, in Table \ref{tab5}, we only present the results for the original MSE loss and the smooth-running loss functions. It can be observed that the improved loss function achieved an accuracy of $90.1\%$.

\section{Conclusion}
\begin{figure}[h]
\centering
\includegraphics[width=\linewidth,height=3in]{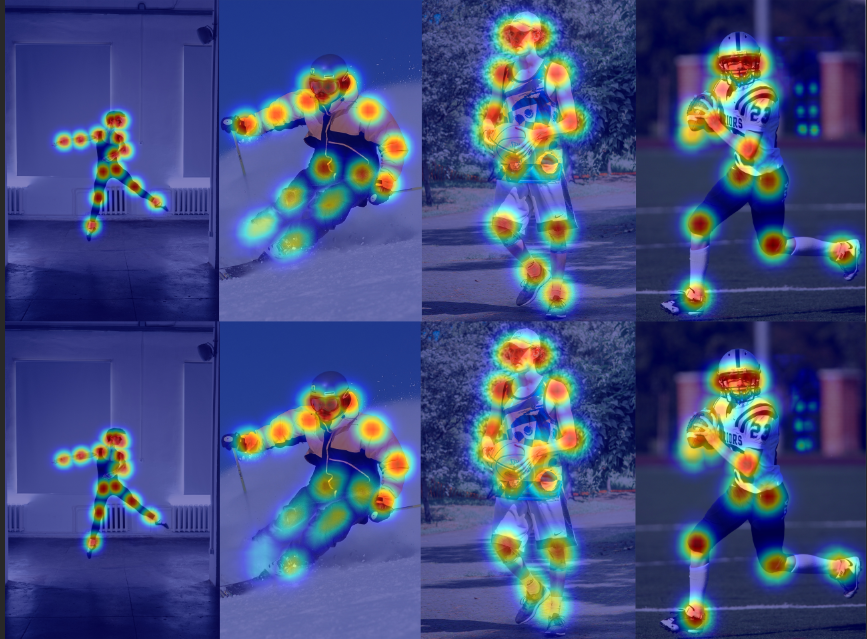}
\caption{Heatmaps generated by different networks}
\label{fig:heatmap}
\end{figure}

In this study, we provide SADI-Net, an innovative framework that combines the RFM, SFM, and DLM. The RFM incorporates a newly designed dilated residual block, which enhances the receptive fields while preserving spatial information perception. The SFM then combines multi-scale features using global and local attention mechanisms from the RFM. With two attention strategies, the model can gain improved details about both local and worldwide features and generate a classification output. Additionally, the DLM learns the distribution of the training data through a flow model based approach to produce trainable heatmaps. Finally, the trainable heatmap will be utilised alongside the RFM-generated heatmap to compute the MSE loss constraint for the weight $\chi$ control. As shown in Figure \ref{fig:heatmap}, we observe that the heatmaps (Upper pictures) inferred by SADI-Net is more accurate in the estimation of difficult critical points than other models (Lower pictures). Our SADI-Net has been shown successful in trials on two popular human pose datasets and achieves superior performance in challenging scenarios, including occlusions, complex backgrounds, and overlapping instances.

\begin{figure}[h]
\centering
\includegraphics[width=\linewidth]{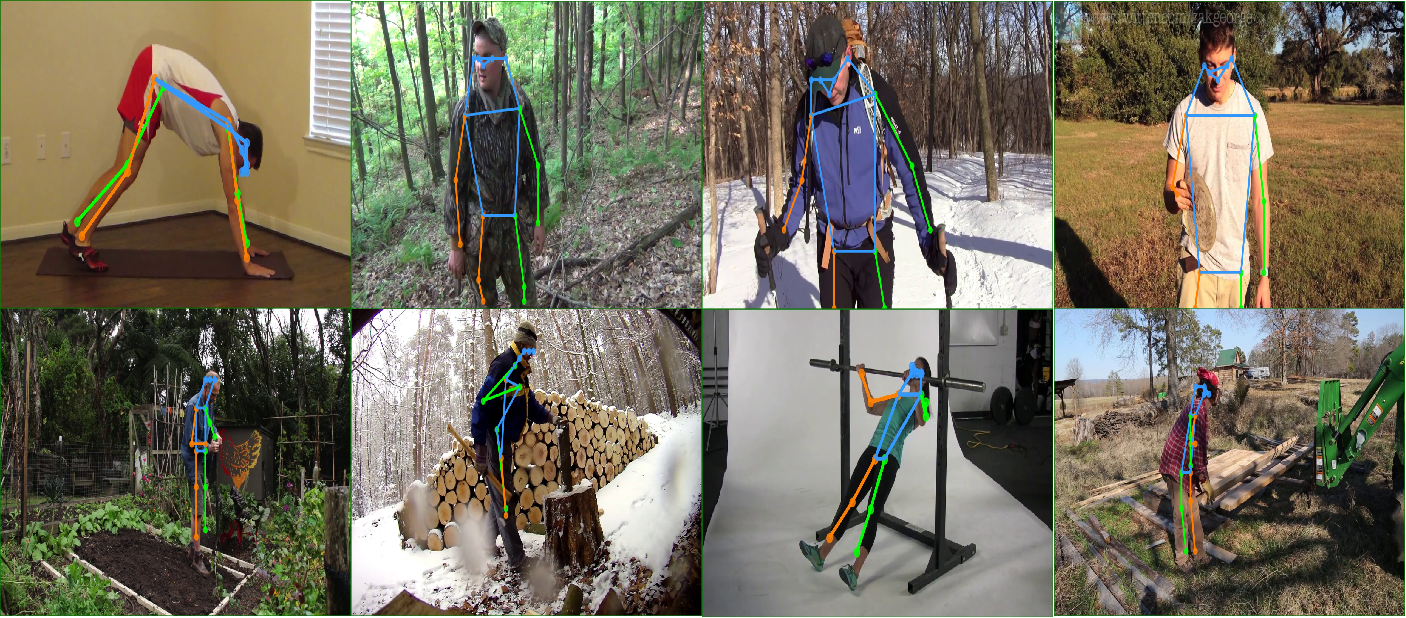}
\caption{Visualisation of the human pose estimation result on the MPII dataset.}
\label{fig:show}
\end{figure}

\section*{Acknowledgment}
Sihan Gao and Jing Zhu contributed equally to this work.

\EOD

\bibliographystyle{plain}
\bibliography{ref/ref.bib}

\begin{IEEEbiography}[{\includegraphics[width=1in,height=1.25in,clip,keepaspectratio]{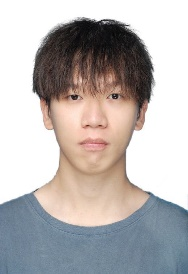}}]{SIHAN GAO} is currently pursuing the B.E. degree with the School of Mathematics and Information Science, Guangzhou University. His research interests include deep learning, human pose estimation, and computer vision.
\end{IEEEbiography}

\begin{IEEEbiography}[{\includegraphics[width=1in,height=1.25in,clip,keepaspectratio]{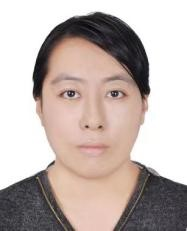}}]{Jing Zhu} Master of Optics, Lecturer. Graduated from the South China Normal University in 2005. Worked in Guangzhou University. Her research interests include mobile robot and image processing technology.
\end{IEEEbiography}

\begin{IEEEbiography}[{\includegraphics[width=1in,height=1.25in,clip,keepaspectratio]{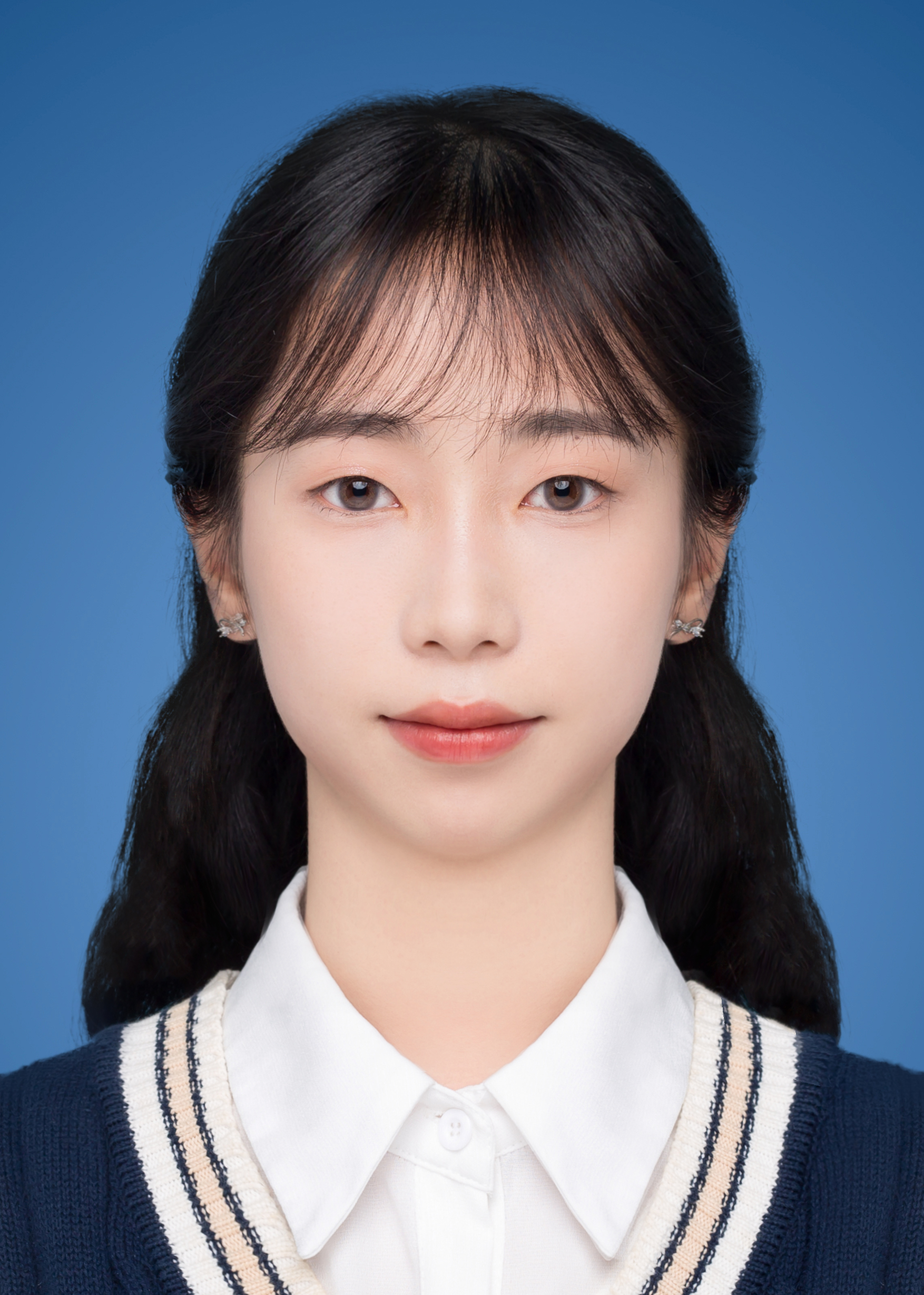}}]{Xiaoxuan Zhuang} (M'87) is currently undergraduating in the School of Electronics and Communication Engineering, Guangzhou University, China. Her research interests include digital image processing, lightweight network design  and analysis in human pose estimation, deep learning, and computer vision. 
\end{IEEEbiography}

\begin{IEEEbiography}[{\includegraphics[width=1in,height=1.25in,clip,keepaspectratio]{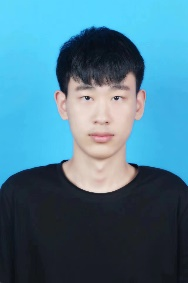}}]{Xiaoxuan Zhuang} is currently pursuing the B.E. degree with the College of Electronic and Information Engineering, Guangzhou University. His research interests include signal processing. design and analysis of algorithm, machine learning and deep learning.
\end{IEEEbiography}

\begin{IEEEbiography}[{\includegraphics[width=1in,height=1.25in,clip,keepaspectratio]{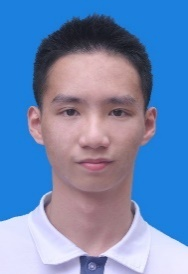}}]{Qijin Li} is currently pursuing the B.E. degree in the Department of Electronic Information Engineering, Guangzhou University, Guangdong, China. His research interests include digital image processing, algorithm design and analysis, deep learning, and computer vision.
\end{IEEEbiography}

\end{document}